\documentclass{article}
\usepackage{arxiv}  % Ensure this is configured correctly
\usepackage[utf8]{inputenc} % Allow UTF-8 input
\usepackage[T1]{fontenc}    % Use 8-bit T1 fonts
\usepackage{hyperref}       % Hyperlinks
\usepackage{url}            % Simple URL typesetting
\usepackage{booktabs}       % Professional-quality tables
\usepackage{amsfonts}       % Blackboard math symbols
\usepackage{nicefrac}       % Compact symbols for 1/2, etc.
\usepackage{microtype}      % Microtypography
\usepackage{lipsum}         % Lorem ipsum (dummy text)
\usepackage{graphicx}       % Graphics inclusion
\usepackage{multirow}       % Multirow tables
\usepackage{siunitx}        % SI units
\usepackage{amsmath}        % Math support
\usepackage{algorithm}      % Algorithm support
\usepackage{algorithmic}    % Algorithm formatting

\graphicspath{{./images/}}  % Ensure images are in ./images/ directory

\title{Leveraging Vision-Language Embeddings for Zero-Shot Learning in Histopathology Images}

\author{
    Md Mamunur Rahaman \\
    School of Computer Science and Engineering\\
    University of New South Wales\\
    Sydney, NSW 2052, Australia \\
    \texttt{md\char`\_mamunur.rahaman@unsw.edu.au} \\
    \And
    Ewan K. A. Millar \\
    Department of Anatomical Pathology\\
    NSW Health Pathology, St. George Hospital\\
    NSW 2217, Australia \\
    \And
    Erik Meijering \\
    School of Computer Science and Engineering\\
    University of New South Wales\\
    Sydney, NSW 2052, Australia \\
}

\begin{document}

\maketitle

\begin{abstract}
Zero-shot learning (ZSL) holds tremendous potential for histopathology image analysis by enabling models to generalize to unseen classes without extensive labeled data. Recent advancements in vision-language models (VLMs) have expanded the capabilities of ZSL, allowing models to perform tasks without task-specific fine-tuning. However, applying VLMs to histopathology presents considerable challenges due to the complexity of histopathological imagery and the nuanced nature of diagnostic tasks. In this paper, we propose a novel framework called Multi-Resolution Prompt-guided Hybrid Embedding (MR-PHE) to address these challenges in zero-shot histopathology image classification. MR-PHE leverages multiresolution patch extraction to mimic the diagnostic workflow of pathologists, capturing both fine-grained cellular details and broader tissue structures critical for accurate diagnosis. We introduce a hybrid embedding strategy that integrates global image embeddings with weighted patch embeddings, effectively combining local and global contextual information. Additionally, we develop a comprehensive prompt generation and selection framework, enriching class descriptions with domain-specific synonyms and clinically relevant features to enhance semantic understanding. A similarity-based patch weighting mechanism assigns attention-like weights to patches based on their relevance to class embeddings, emphasizing diagnostically important regions during classification. Our approach utilizes pretrained VLM, CONCH for ZSL without requiring domain-specific fine-tuning, offering scalability and reducing dependence on large annotated datasets. Experimental results demonstrate that MR-PHE not only significantly improves zero-shot classification performance on histopathology datasets but also often surpasses fully supervised models, highlighting its superior effectiveness and potential for advancing computational pathology.
\end{abstract}

%\begin{keywords} 
%Vision-Language Models (VLMs), Histopathology, Zero-Shot Learning, Hybrid Embedding, Prompt Generation, Computational Pathology.
%\end{keywords}

\section{Introduction}
\label{sec:introduction}

Recent advancements in vision-language models (VLMs) have markedly expanded the scope of vision-based systems, enabling them to comprehend open vocabularies and perform zero-shot learning (ZSL) across diverse domains \cite{du2022learning, li2023citetracker, shao2023prompting, zhang2023survey, zhang2021vinvl, zhou2022learning, clip2021radford}. Models such as CLIP \cite{clip2021radford} and ALIGN \cite{align2021jia} have demonstrated substantial potential by aligning images with textual descriptions through extensive pretraining on large-scale datasets. This alignment facilitates various tasks, including image classification, retrieval, and captioning, without the need for task-specific fine-tuning. However, applying VLMs in specialized domains such as histopathology poses substantial challenges due to the complexity of histopathological imagery and the intricate nature of diagnostic tasks. These challenges necessitate novel approaches to adapt VLMs effectively for histopathology.

Histopathological image analysis remains the cornerstone of cancer diagnosis, typically relying on microscopic examination of tissue slides. However, evaluations by pathologists can be subjective, leading to variability in diagnostic accuracy. Reported accuracy rates for breast cancer subtyping and prostate cancer grading, for instance, have been as low as 57.08\% and agreement rates around 57.90\%~\cite{pati2022hierarchical, yang2012does}. In response, there has been an increasing trend toward digitization of this process using artificial intelligence (AI) and computer-aided diagnosis (CAD) systems. Convolutional neural networks (CNNs), in particular, have demonstrated considerable potential in autonomously extracting key features from histopathological images, improving diagnostic reliability and reducing inter-observer variability. Over the past decade, CNN models have achieved notable advancements in analyzing histopathological images, contributing to more accurate detection and diagnosis across various cancer types. These models effectively extract complex features from raw images, enhancing diagnostic precision \cite{Rahaman2024Histopathology, li2019classification}. Notable developments include the integration of transfer learning techniques \cite{cao2018improve, xie2019deep}, which leverage pretrained models to improve performance on domain-specific tasks, and ensemble learning strategies \cite{karthik2022classification}, which enhance robustness by combining multiple models. Additionally, innovations in algorithmic design and architecture \cite{pati2022hierarchical, brancati2022bracs, tiard2022stain, wang2022transformer, Riasatian2021FineTuning, hu2023ebhi, chen2022gashis, fan2023cam} have optimized image classification workflows, resulting in more accurate and efficient analyses.

However, despite these advancements, supervised learning approaches still face substantial limitations. They require large, annotated datasets, which are often expensive and time-consuming to compile, and they are prone to overfitting when trained on homogeneous data \cite{rahaman2024generalized}. Furthermore, supervised models often underperform when applied to unseen datasets, and their deployment in clinical environments is restricted by their inability to generalize across diverse imaging modalities and histopathological domains. These limitations highlight the necessity for alternative approaches that can overcome the reliance on extensive labeled datasets and improve generalization to unseen data. ZSL emerges as a promising solution, enabling models to recognize and classify unseen classes without requiring domain-specific training examples. By leveraging semantic information and facilitating knowledge transfer, ZSL offers enhanced scalability, reducing the dependence on large annotated datasets and improving the adaptability of models to dynamic and heterogeneous medical data. These advantages make ZSL a pivotal advancement for scalable and generalized histopathological image analysis systems.

%In computational pathology, VLMs have shown considerable promise for histopathological image analysis, particularly for tasks such as tissue recognition and cancer subtyping. Lu et al.~\cite{lu2023visual} fine-tuned CLIP on 33.48k histology image-caption pairs for cancer subtyping, while Huang et al.~\cite{huang2023visual} enhanced zero-shot classification by leveraging over 200k Medical Twitter histology images and textual data. Zhang et al.~\cite{zhang2023large} expanded the capabilities of VLMs by training BiomedCLIP on 15 million biomedical image-text pairs. More recently, Lu et al.~\cite{lu2024visual} introduced CONCH, a foundation model pretrained on 1.17 million histopathology image-caption pairs, for improved performance in various histopathological tasks. These advancements highlight the growing significance of VLMs in advancing ZSL and automated pathology systems, though challenges remain in aligning complex cancer morphologies with textual representations.

In computational pathology, VLMs have shown considerable promise for histopathological image analysis, particularly in tasks like tissue recognition and cancer subtyping. Recent studies have adapted models such as CLIP and developed new VLMs by training on large-scale biomedical image-text datasets to enhance zero-shot classification and expand capabilities in the domain \cite{lu2023visual, huang2023visual, zhang2023large, lu2024visual}. These advancements highlight the growing significance of VLMs in advancing ZSL and automated pathology systems, though challenges remain in aligning complex cancer morphologies with textual representations.

Despite the progress in VLMs for computational pathology, current systems are constrained by several limitations. Most notably, aligning complex and diverse cancer morphologies with simplistic, single-phrase textual descriptions remains a challenge. %Existing methods often rely on noun-based terms, which fail to capture the nuanced pathological features necessary for accurate classification, leading to suboptimal results.
Existing ZSL methods depend on simplistic, noun-based textual prompts (e.g., using only class names) when applied to histopathology. Such prompts are insufficient to capture the complex, nuanced morphological features needed for accurate diagnosis. Furthermore, generating effective prompts for VLMs is a labor-intensive process that requires domain expertise, limiting scalability. While ZSL has been extensively explored in natural image domains \cite{clip2021radford, menon2022visual, pratt2023what, roth2023waffling, li2024visual} and in certain medical imaging modalities, such as radiology \cite{tiu2022expert, wang2022medclip, zhang2022contrastive}, there remains a lack of contributions focused on ZSL for histopathology. Given the challenges associated with training large supervised models in data-scarce domains, ZSL presents an effective alternative. By enabling models to generalize without extensive domain-specific training, ZSL offers a scalable solution, especially in medical fields where annotated data is often limited.
%This positions ZSL as the next major advancement for computational pathology.

\begin{figure}[!t]
\centering
\includegraphics[scale=0.65]{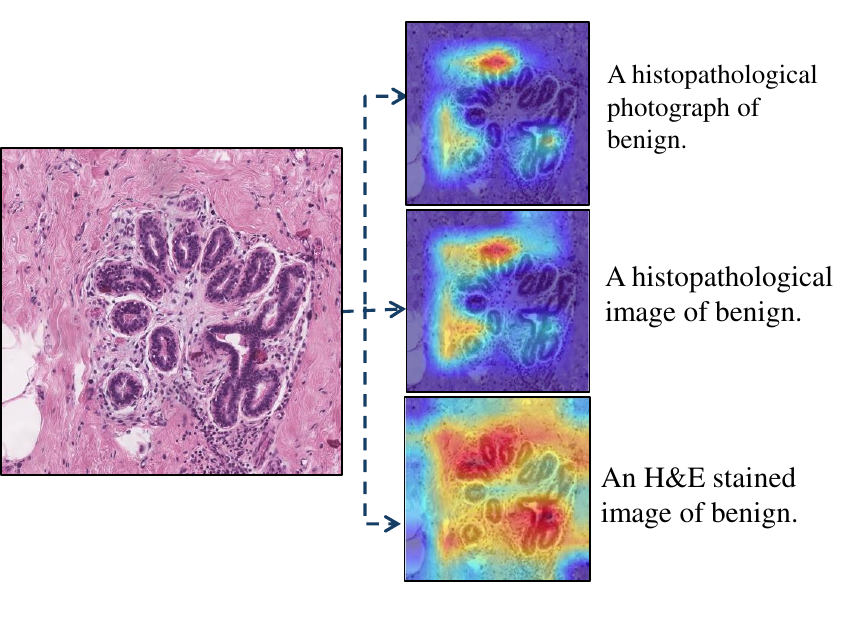} 
\caption{An H\&E-stained histopathological image of benign breast tissue from the BRACS dataset (resolution: 2472~$\times$~2370 pixels). The Grad-CAM visualization highlights regions of interest identified by the VLM CONCH model. Warmer colors in the heatmap indicate areas where the model assigns higher attention weights, corresponding to features relevant for benign classification.}
\label{fig:prompt_attention}
%\vspace{-7.5mm}
\end{figure}

In ZSL, different text prompts can direct attention to varying regions within the same histopathological image (Fig.~\ref{fig:prompt_attention}). Finer-grained descriptions, such as those specifying malignant regions, may result in localized alignment with particular areas of the image but might not fully capture the broader context of the entire tissue sample. This challenge underscores the difficulty of aligning detailed textual descriptions with complex histopathology images, where both localized features and global tissue context are crucial for accurate diagnosis.

To address these challenges, we propose a novel framework called Multi-Resolution Prompt-guided Hybrid Embedding (MR-PHE) for ZSL in histopathology image classification. Our method mimics the diagnostic workflow of pathologists by implementing a multiresolution patch extraction technique, analyzing images at varying scales to capture both fine-grained cellular details and broader tissue structures critical for accurate diagnosis. We introduce a hybrid embedding strategy that integrates global image embeddings with weighted patch embeddings derived from these multiresolution patches. A similarity-based patch weighting mechanism assigns attention-like weights to patches based on their relevance to class embeddings, emphasizing diagnostically important regions while maintaining a holistic understanding of the entire image. Additionally, we develop a comprehensive prompt generation and selection framework, generating domain-specific textual prompts enriched with synonyms and clinically relevant features to enhance semantic understanding. By filtering and selecting the most effective prompts for each class, our method ensures that both critical localized regions and the overall tissue structure are accurately captured and aligned with their corresponding descriptions. This leads to more robust and nuanced classifications, allowing the model to focus on diagnostically relevant regions without losing context. Our approach leverages VLMs for ZSL without requiring domain-specific fine-tuning, offering scalability and reducing dependence on large annotated datasets. 

Experimental evaluations on multiple histopathology datasets demonstrate that the proposed MR-PHE not only significantly improves zero-shot classification performance but also often surpasses fully supervised models, validating the effectiveness of our approach in real-world applications. These results underscore the potential of MR-PHE to facilitate histopathological image analysis without the need for extensive labeled data, advancing the capabilities of computational pathology.

\section{Literature Review}
\label{sec:review}
Recent advances in machine learning have led to sophisticated models for computational pathology, particularly for analyzing histopathological images. Key strategies include weakly-supervised learning (WSL), self-supervised learning (SSL), and VLMs, each addressing challenges like limited labeled data, tissue variability, and the complexity of accurate diagnosis. This section reviews these approaches, their strengths and limitations, and situates our proposed method within this evolving field.

\subsection{Weakly-Supervised Learning in Histopathology}
WSL approaches in histopathology have emerged as crucial methods to mitigate the challenges of acquiring detailed annotations for histopathological images. Multiple instance learning (MIL) \cite{amores2013multiple} frameworks, in particular, have been extensively applied to whole slide images (WSIs), enabling models to operate with image-level labels instead of requiring precise pixel-wise annotations. Notable methods such as ABMIL \cite{ilse2018attention} and CLAM \cite{lu2021data} have made significant contributions to WSI classification by learning bag-level representations through weak labels. For instance, CLAM employs attention mechanisms to focus on diagnostically significant regions in the image. These MIL-based approaches, while effective, still rely on the availability of labeled data at a high level and may struggle with the fine-grained details critical for accurate diagnosis \cite{javed2022additive, NEURIPS2022wang, xiang2023exploring, cui2023bayes, zhang2022dtfd}. Our proposed method, while benefiting from these insights, diverges from WSL by leveraging ZSL and VLMs to bypass the dependency on labeled training data entirely, thus addressing scalability and adaptability concerns in data-scarce domains like histopathology.

\subsection{Self-Supervised Learning in Histopathology}
SSL has become a key technique in histopathology for deriving meaningful representations from large, unlabeled datasets. Contrastive learning, a widely used SSL approach, trains models by distinguishing between different augmentations of the same image and contrasting them with other images. Recent studies in computational pathology have applied contrastive learning to unlabeled histopathology datasets, enabling the pretraining of domain-specific encoders, which show notable performance improvements on downstream tasks \cite{chen2022scaling, wang2022transformer, an2022masked, ciga2022self, koohbanani2021self, srinidhi2021improving, vu2023handcrafted}. However, SSL often requires extensive data augmentation and auxiliary tasks, which can limit its scalability and applicability in complex medical imaging tasks. Our approach overcomes these limitations by integrating cross-modal learning with VLMs, achieving enhanced zero-shot performance without reliance on augmentations or auxiliary techniques.

\subsection{Vision-Language Models for Computational Pathology}
VLMs have recently been introduced in computational pathology to bridge the gap between visual data and medical textual information. Models like BiomedCLIP \cite{zhang2023large}, MI-Zero \cite{lu2023visual}, and PLIP \cite{huang2023visual} have adapted the CLIP \cite{clip2021radford} architecture for aligning pathology images with textual descriptions, while CONCH \cite{lu2023towards} is based on the CoCa \cite{yu2022coca} framework, which combines contrastive and captioning losses to achieve similar alignment. These models are typically trained on large-scale datasets like OpenPath \cite{huang2023visual} and Quilt-1M \cite{ikezogwo2024quilt}, which contain millions of image-text pairs and require extensive computational resources to train from scratch. However, once pretrained, these foundation models can be leveraged for downstream tasks such as ZSL without the need for further training, making them highly adaptable and scalable for new applications. In both the natural image domain \cite{clip2021radford, menon2022visual, pratt2023what, roth2023waffling, li2024visual} and medical imaging \cite{tiu2022expert, wang2022medclip, zhang2022contrastive}, considerable research has focused on using these pretrained models for ZSL, which circumvents the need for labeled data while still achieving strong performance. Despite this progress, a key limitation of current VLMs is their reliance on single-text and single-image alignment strategies, which may fail to capture the complex, multimodal relationships in pathology. Our approach, through hybrid embeddings and prompt-based alignment, builds upon these VLMs by aligning multiple correlated visual and textual concepts, enabling more robust generalization across a broader spectrum of pathology tasks.

\begin{figure*}[!t]
\centering
\includegraphics[scale=0.55]{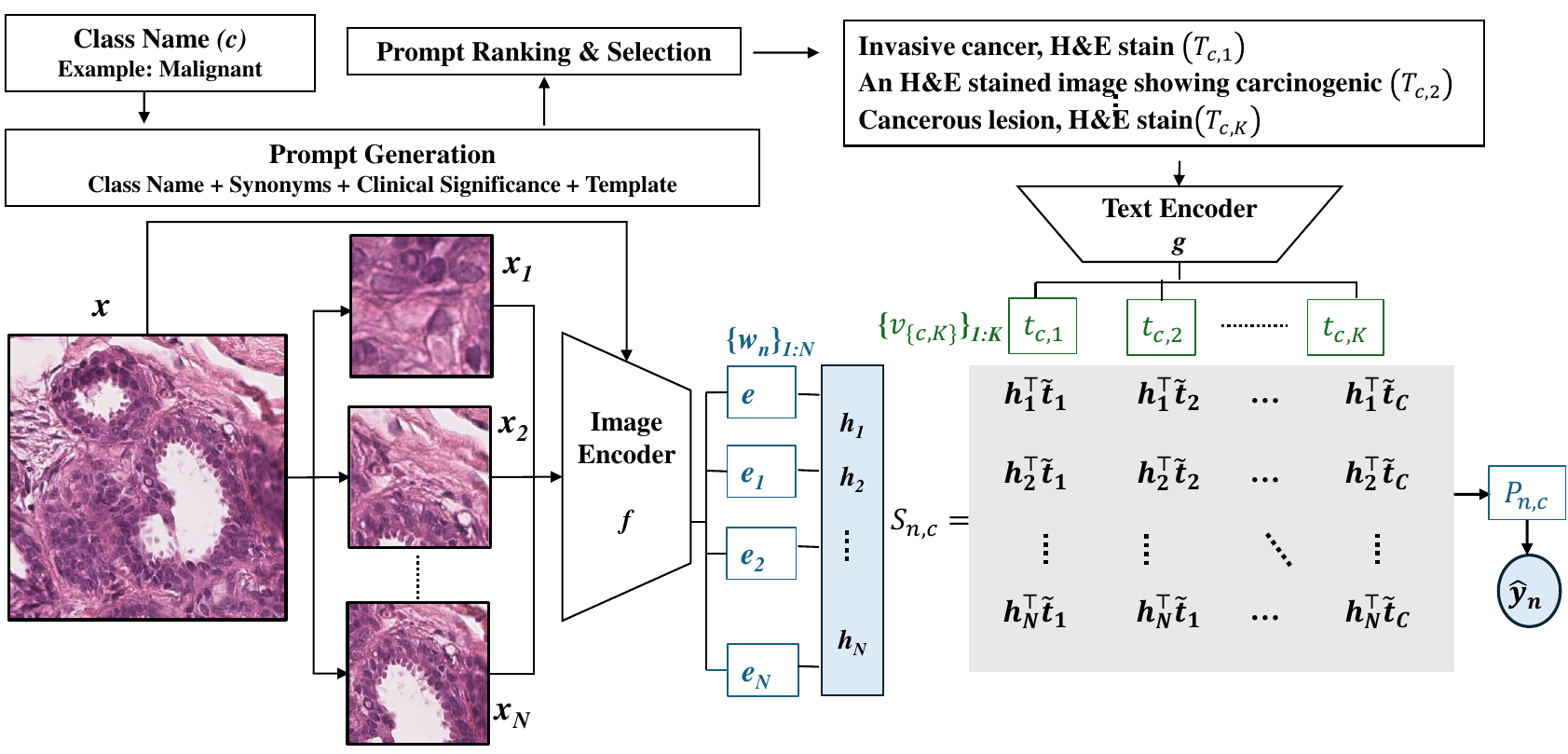} 

\caption{Workflow of our proposed MR-PHE framework. The input image \( x \) is divided into multiresolution patches \( \{x_1, x_2, \dots, x_N\} \), which are encoded into embeddings \( \{e_1, e_2, \dots, e_N\} \) via the image encoder \( f \). A hybrid embedding \( h \) is created by combining global and weighted patch-level features using attention weights \( \{w_1, w_2, \dots, w_N\} \). Simultaneously, class-specific textual prompts \( \{T_{c,1}, T_{c,2}, \dots, T_{c,K}\} \) are generated for each class \( c \), ranked, and encoded by the text encoder \( g \) into prompt embeddings \( \{t_{c,1}, t_{c,2}, \dots, t_{c,K}\} \). Text weights \( \{v_{c,1}, v_{c,2}, \dots, v_{c,K}\} \) are computed to aggregate these embeddings into the final class embeddings \( \tilde{t}_c \). The similarity scores \( S_{n,c} = h^\top \tilde{t}_c \) are computed between the hybrid image embedding \( h \) and each class embedding \( \tilde{t}_c \). These scores \( S_{n,c} \) are scaled and converted into class probabilities \( P_{n,c} \) using the softmax function. The predicted class label \( \hat{y}_n \) is then determined by selecting the class with the highest probability \( P_{n,c} \).}
\label{fig:2}
\vspace{-5mm}
\end{figure*}

\section{Methodology} 
\label{sec:Method}

Our proposed MR-PHE framework (Fig.~\ref{fig:2}) leverages a combination of multiresolution patch extraction, hybrid embeddings, and prompt-guided learning to capture both fine-grained and global features of histopathology images. This section presents the key components of the framework (see Algorithm~\ref{alg:MR-PHE} for a detailed technical summary of all steps).

\subsection{Multiresolution Patch Extraction and Image Embedding}

Given an input image \( x \), we extract a set of patches \( \{ x_1, x_2, \dots, x_N \} \) at multiple resolutions. A set of scaling factors \( \mathcal{S} = \{ s_1, s_2, \dots, s_S \} \), where each scaling factor \( s_k \in (0, 1] \), represents a series of resolution levels. In our implementation, we use scaling factors \( \mathcal{S} = \{0.25, 0.50, 0.75\} \), corresponding to 25\%, 50\%, and 75\% of the original image size. For each scaling factor \( s_k \), the input image \( x \) is resized to obtain a scaled image \( x^{(s_k)} \):
\begin{equation}
x^{(s_k)} = \text{Resize}(x, s_k).
\end{equation}
From each scaled image \( x^{(s_k)} \), we extract \( n_k \) random patches using a cropping function \( \mathcal{C} \):
\begin{equation}
x_{i}^{(s_k)} = \mathcal{C}(x^{(s_k)}), \quad i = 1, 2, \dots, n_k.
\end{equation}
Here, \( \mathcal{C} \) dynamically selects a crop size within a predefined range proportional to the scaled image dimensions. The cropping process is randomized in terms of location and size to ensure diverse sampling of features, including localized details and broader tissue structures. The total number of patches is \( N = \sum_{k=1}^{S} n_k \), and in our experiments, we extract \( n_k = 5 \) patches per scale, resulting in \( N = 15 \) patches in total. The complete set of patches is:
\begin{equation}
\{ x_1, x_2, \dots, x_N \} = \bigcup_{k=1}^{S} \{ x_{1}^{(s_k)}, x_{2}^{(s_k)}, \dots, x_{n_k}^{(s_k)} \}.
\end{equation}
We retain the original image \( x \) to capture global contextual information, resulting in \( N+1 \) images for embedding.

\subsubsection{Image Embedding}
We utilize a pretrained image encoder \( f \), specifically the CONCH model~\cite{lu2024visual}, to encode the patches and the original image into embeddings. Each patch \( x_i \) is encoded to obtain its corresponding embedding \( e_i \):
\begin{equation}
e_i = f(x_i), \quad i = 1, 2, \dots, N.
\end{equation}
The original image \( x \) is encoded to obtain the global embedding \( e_{\text{global}} \):
\begin{equation}
e_{\text{global}} = f(x).
\end{equation}
All embeddings are normalized to unit vectors for efficient cosine similarity computations:
\begin{equation}
\tilde{e}_i = \frac{e_i}{\| e_i \|}, \quad \tilde{e}_{\text{global}} = \frac{e_{\text{global}}}{\| e_{\text{global}} \|}.
\end{equation}

\subsection{Hybrid Embedding Construction}\label{sec:hybrid_embedding}
To integrate the global context with local details, we construct a hybrid embedding \( h \) by combining the global embedding \( \tilde{e}_{\text{global}} \) with the weighted patch embeddings. 

\subsubsection{Attention Weights for Patch Embeddings}
For each patch embedding \( \tilde{e}_i \), attention weights \( w_i \) are computed based on their cosine similarity to the class embeddings. Let \( \{ t_1, t_2, \dots, t_C \} \) denote the class embeddings, where the \( t_c \) are normalized. For each patch \( \tilde{e}_i \), the similarity score with each class embedding \( t_c \) is computed as:
\begin{equation}
s_{i,c} = \tilde{e}_i^\top t_c.
\end{equation}
We compute the maximum similarity for each patch across all classes:
\begin{equation}
s_i = \max_{c} s_{i,c}.
\end{equation}
The attention weight \( w_i \) is then computed using a softmax function over the maximum similarity scores:
\begin{equation}
w_i = \frac{\exp(s_i)}{\sum_{j=1}^{N} \exp(s_j)}.
\end{equation}

\subsubsection{Aggregating Patch Embeddings}
The weighted patch embedding \( e_{\text{patch}} \) is then computed as:
\begin{equation}
e_{\text{patch}} = \sum_{i=1}^{N} w_i \tilde{e}_i.
\end{equation}

\subsubsection{Forming the Hybrid Embedding}
The hybrid embedding \( h \) is constructed by combining the global embedding and the weighted patch embedding:
\begin{equation}
h = \alpha \tilde{e}_{\text{global}} + (1 - \alpha) e_{\text{patch}}
\end{equation}
where \( \alpha \) is a weighting factor that balances the contribution of global and local features. %In our implementation, \( \alpha = 0.6 \).

\subsection{Text Encoder and Prompt-Based Class Embedding}\label{sec:class_embedding}
To effectively capture the semantic richness and clinical significance of each class in histopathology image classification, we employ a prompt-based approach for generating diverse textual descriptions. These prompts are encoded using a pretrained text encoder \( g \) (from the CONCH model~\cite{lu2024visual}) to obtain robust class embeddings \( \{ t_1, t_2, \dots, t_C \} \), where \( C \) is the number of classes.

\subsubsection{Prompt Generation}
For each class \( c \), we generate a comprehensive set of textual prompts \( \mathcal{T}_c = \{ T_{c,1}, T_{c,2}, \dots, T_{c,M_c} \} \) by combining:
\begin{itemize}
    \item \textbf{Synonyms and Descriptors} (\( \mathcal{S}_c \)): Alternative names and descriptors relevant to class \( c \), capturing various terminologies used in clinical practice. We employ GPT-4 to suggest candidate synonyms and descriptors, which are then screened to ensure domain relevance.
    \item \textbf{Templates} (\( \mathcal{P} \)): Predefined sentence structures with place\-holders for class names or synonyms.
    \item \textbf{Clinical Significance Statements} (\( \mathcal{C}_c \)): Detailed descriptions highlighting important clinical and pathological features associated with class \( c \). These statements are also initially generated via GPT-4 and curated to preserve accuracy.
\end{itemize}
Formally, the set of prompts for class \( c \) is constructed as:
\begin{equation}
\begin{aligned}
\mathcal{T}_c = \{ & \text{template}(\text{synonym}) \mid \\
                & \text{synonym} \in \mathcal{S}_c, \ \text{template} \in \mathcal{P} \} \cup \mathcal{C}_c.
\end{aligned}
\end{equation}
This approach ensures that the prompts encapsulate both general and specific characteristics of each class.

\subsubsection{Text Embedding Computation}
Each prompt \( T_{c,m} \in \mathcal{T}_c \) is encoded using the text encoder \( g \) (pretrained from the CONCH model) to obtain its embedding:
\begin{equation}
t_{c,m} = g(T_{c,m}), \quad m = 1, 2, \dots, M_c.
\end{equation}
The embeddings are normalized to unit vectors to facilitate cosine similarity computations:
\begin{equation}
\tilde{t}_{c,m} = \frac{t_{c,m}}{\| t_{c,m} \|}.
\end{equation}

\subsubsection{Prompt Evaluation and Selection}
To identify the most effective prompts, we evaluate each prompt \( T_{c,m} \) based on its classification performance. For each prompt, we perform the following steps:
\begin{enumerate}
    \item \textbf{Compute Similarities}: Calculate the cosine similarity between the normal image embedding \( e_n \) (obtained directly from the image encoder without hybrid embedding) and the prompt embedding \( \tilde{t}_{c,m} \):
    \begin{equation}
    s_{n,m} = e_n^\top \tilde{t}_{c,m}, \quad n = 1, 2, \dots, N_{\text{val}}.
    \end{equation}
    where \( N_{\text{val}} \) is the number of validation samples.
    \item \textbf{Predict Labels}: For each image \( n \), predict the class label \( \hat{y}_n \) by selecting the class \( c \) that maximizes the similarity score across all prompts for all classes:
    \begin{equation}
    \hat{y}_n = \arg\max_{c} \max_m s_{n,m}, \quad c = 1, 2, \dots, C.
    \end{equation}
    \item \textbf{Compute Accuracy}: Calculate the accuracy of the prompt \( T_{c,m} \) by comparing the predicted labels with the ground truth.
\end{enumerate}
After evaluating all prompts, we select the top \( K \) prompts \( \mathcal{T}_c^{\text{top}} \subseteq \mathcal{T}_c \) for each class \( c \) based on their classification accuracy.

\subsubsection{Class Embedding Construction}
The final class embedding \( t_c \) for class \( c \) is obtained by averaging the embeddings of its top \( K \) prompts:
\begin{equation}
t_c = \frac{1}{K} \sum_{T_{c,m} \in \mathcal{T}_c^{\text{top}}} \tilde{t}_{c,m}.
\end{equation}
We then normalize the class embedding:
\begin{equation}
\tilde{t}_c = \frac{t_c}{\| t_c \|}.
\end{equation}
This aggregated embedding \( \tilde{t}_c \) captures the most salient textual features relevant to class \( c \), providing a robust representation for subsequent classification tasks.

\subsection{Similarity Matrix Calculation and Final Prediction}
\label{sec:similarity_classification}
With the hybrid image embeddings \( \{ h_n \} \) and the class embeddings \( \{ \tilde{t}_c \} \) obtained from the previous steps, we proceed to compute the similarity matrix, apply weights, and make the final prediction.

\subsubsection{Weighted Class Embedding Construction}

To enhance the class embeddings' representational capacity, we refine the selected prompts by weighting based on their relevance to a reference label embedding. This introduces text weights \( v_j \), which emphasize more relevant prompts for each class.

\paragraph{Reference Label Embedding}
For each class \( c \), we define a reference label embedding \( l_c \) as the embedding of a simple label prompt, such as:
\begin{equation}
l_c = g(\text{``a photo of a } \text{CLASSNAME}\text{.''})
\end{equation}
The embedding is normalized to ensure that the following similarity computations are based on unit vectors:
\begin{equation}
\tilde{l}_c = \frac{l_c}{\| l_c \|}.
\end{equation}

\paragraph{Computing Text Weights}
For each prompt embedding \( \tilde{t}_{c,j} \) (from the top \( K \) prompts selected earlier), we compute a similarity score between the prompt embedding and the reference label embedding \( \tilde{l}_c \):
\begin{equation}
s_j = \tilde{t}_{c,j}^\top \tilde{l}_c.
\end{equation}
The text weights \( v_j \) are then computed using a softmax function scaled by a factor \( \beta \) to control the sharpness of the weight distribution:
\begin{equation}
v_j = \frac{\exp(\beta s_j)}{\sum_{k=1}^{K} \exp(\beta s_k)}.
\end{equation}
Here, \( \beta \) is a scaling parameter that determines how strongly the weight distribution is influenced by the similarities.

\paragraph{Weighted Class Embedding}
The final class embedding \( \tilde{t}_c \) is constructed as a weighted sum of the top \( K \) prompt embeddings:
\begin{equation}
\tilde{t}_c = \sum_{j=1}^{K} v_j \tilde{t}_{c,j}.
\end{equation}
This aggregation emphasizes the most relevant prompts for each class, providing a more robust class representation.

\subsubsection{Weighted Cosine Similarity Calculation}
Next, we compute the cosine similarity between each hybrid image embedding \( h_n \) and the final weighted class embeddings \( \tilde{t}_c \). Given that both embeddings are normalized, the cosine similarity reduces to the dot product:
\begin{equation}
S_{n,c} = h_n^\top \tilde{t}_c
\label{eq:Snc}
\end{equation}
resulting in similarity matrix \( S \in \mathbb{R}^{N \times C} \), where \( N \) and \( C \) are the numbers of images and classes. The matrix contains the similarity scores between each image and each class.

\subsubsection{Temperature Scaling}
To control the confidence in the similarity scores, we apply temperature scaling, which sharpens or smooths the probability distribution:
\begin{equation}
\hat{S}_{n,c} = \frac{S_{n,c}}{\tau}
\end{equation}
where \( \tau > 0 \) is the temperature parameter. A higher (lower) \( \tau \) produces softer (sharper) class probability distributions. %In our implementation, we use \( \tau = 1.50 \).

\subsubsection{Probability Computation and Classification}
The scaled similarity scores \( \hat{S}_{n,c} \) are converted into class probabilities using the softmax function:
\begin{equation}
P_{n,c} = \frac{\exp(\hat{S}_{n,c})}{\sum_{k=1}^{C} \exp(\hat{S}_{n,k})}.
\end{equation}
Finally, the predicted class label \( \hat{y}_n \) for each image \( n \) is determined by selecting the class with the highest probability:
\begin{equation}
\hat{y}_n = \arg\max_{c} P_{n,c}.
\end{equation}
This ensures that each image is classified based on the class with the highest weighted similarity to its hybrid embedding, making the final prediction.
\begin{algorithm}[!ht]
\caption{Multi-Resolution Prompt-guided Hybrid Embedding (MR-PHE)}
\label{alg:MR-PHE}
\begin{algorithmic}[1]
\footnotesize
\STATE \textbf{Input:}
\begin{itemize}
    \item Image $x$
    \item Class labels $Y = \{ c \}_{c=1}^C$
    \item Pretrained image encoder $f$
    \item Pretrained text encoder $g$
    \item Scaling factors $\mathcal{S} = \{ s_k \}_{k=1}^S$
    \item Number of patches per scale $\{ n_k \}_{k=1}^S$
    \item Hybrid weight $\alpha$
    \item Temperature $\tau$
    \item Text scaling factor $\beta$
    \item Number of top prompts $K$
    \item Random cropping function $\mathcal{C}$
\end{itemize}
\STATE \textbf{Output:} Predicted class label $\hat{y}$
\STATE Initialize the patch set $\mathcal{P} \leftarrow \emptyset$
\FOR{$k = 1$ to $S$}
    \STATE Resize image: $x^{(s_k)} \leftarrow \text{Resize}(x, s_k)$
    \FOR{$i = 1$ to $n_k$}
        \STATE Obtain patch: $x_{k,i} \leftarrow \mathcal{C}(x^{(s_k)})$
        \STATE Add $x_{k,i}$ to $\mathcal{P}$
    \ENDFOR
\ENDFOR
\STATE Include original image $x$ in $\mathcal{P}$
\FOR{each image $x_p \in \mathcal{P}$}
    \STATE Compute embedding: $e_p \leftarrow f(x_p)$
    \STATE Normalize: $\tilde{e}_p \leftarrow e_p / \| e_p \|$
\ENDFOR
\STATE Let $\tilde{e}_{\text{global}}$ be the normalized embedding of $x$
\FOR{each class $c \in Y$}
    \STATE Generate prompts: $\mathcal{T}_c = \{ T_{c,m} \}_{m=1}^{M_c}$
    \FOR{$m = 1$ to $M_c$}
        \STATE Compute embedding: $t_{c,m} \leftarrow g(T_{c,m})$
        \STATE Normalize: $\tilde{t}_{c,m} \leftarrow t_{c,m} / \| t_{c,m} \|$
    \ENDFOR
    \STATE Evaluate prompts $\mathcal{T}_c$ on validation set and select top $K$ prompts: $\mathcal{T}_c^{\text{top}}$
    \STATE Reference label embedding: $l_c \leftarrow g($``a photo of a $c$.'')
    \STATE Normalize: $\tilde{l}_c \leftarrow l_c / \| l_c \|$
    \FOR{each $\tilde{t}_{c,j} \in \mathcal{T}_c^{\text{top}}$}
        \STATE Compute similarity: $s_j \leftarrow \tilde{t}_{c,j}^\top \tilde{l}_c$
    \ENDFOR
    \STATE Compute text weights: $v_j \leftarrow \dfrac{\exp(\beta s_j)}{\sum_{k=1}^{K} \exp(\beta s_k)}$
    \STATE Compute weighted class embedding: $\tilde{t}_c \leftarrow \sum_{j=1}^{K} v_j \tilde{t}_{c,j}$
\ENDFOR
\FOR{each patch embedding $\tilde{e}_p$}
    \FOR{$c = 1$ to $C$}
        \STATE Compute similarity: $s_{p,c} \leftarrow \tilde{e}_p^\top \tilde{t}_c$
    \ENDFOR
    \STATE Compute maximum similarity: $s_p \leftarrow \max_{c} s_{p,c}$
\ENDFOR
\STATE Compute attention weights: $w_p \leftarrow \dfrac{\exp(s_p)}{\sum_{q} \exp(s_q)}$
\STATE Compute weighted patch embedding: $e_{\text{patch}} \leftarrow \sum_{p} w_p \tilde{e}_p$
\STATE Compute hybrid embedding: $h \leftarrow \alpha \tilde{e}_{\text{global}} + (1 - \alpha) e_{\text{patch}}$
\STATE Normalize: $h \leftarrow h / \| h \|$
\FOR{$c = 1$ to $C$}
    \STATE Compute similarity: $S_c \leftarrow h^\top \tilde{t}_c$
\ENDFOR
\STATE Apply temperature scaling: $\hat{S}_c \leftarrow \dfrac{S_c}{\tau}$
\STATE Compute probabilities: $P_c \leftarrow \dfrac{\exp(\hat{S}_c)}{\sum_{k=1}^C \exp(\hat{S}_k)}$
\STATE Predict class label: $\hat{y} \leftarrow \arg\max_{c} P_c$
\STATE \textbf{Return} $\hat{y}$
\end{algorithmic}
\end{algorithm}
\subsection{Zero-Shot Learning for Classification}

ZSL enables models to classify instances of unseen classes by leveraging semantic information, such as textual descriptions, without requiring labeled training data for those classes \cite{clip2021radford}. In our MR-PHE framework, we utilize ZSL by employing pretrained VLMs to align visual features of histopathology images with textual embeddings of class descriptions.

We use pretrained image and text encoders \( f \) and \( g \) from the CONCH model~\cite{lu2024visual}. These encoders have been trained on large-scale datasets containing diverse image-text pairs, allowing them to capture rich visual and semantic representations. By using these encoders without domain-specific fine-tuning, our framework operates in a zero-shot setting, effectively generalizing to unseen histopathology classes.

Our method aligns the visual embeddings of images with the textual embeddings of class descriptions in a shared embedding space. The hybrid image embeddings \( h_n \) (Section~\ref{sec:hybrid_embedding}) capture both local details and global context, enhancing the model's ability to recognize complex patterns in histopathology images. The class embeddings \( \tilde{t}_c \), derived from diverse and clinically relevant prompts (Section~\ref{sec:class_embedding}), represent the semantic information of each class.

The zero-shot classification is performed by computing the cosine similarity between the hybrid image embeddings and the class embeddings as per (\ref{eq:Snc}).
%\begin{equation}
%S_{n,c} = h_n^\top \tilde{t}_c.
%\end{equation}
By converting these similarities into probabilities and selecting the class with the highest probability, we classify images into unseen classes without any prior training on the specific classification task. This approach leverages the generalized knowledge encoded in the pretrained models and enhances it through our hybrid embedding strategy, enabling effective ZSL in histopathology image classification.

%\subsection{Algorithm Summary}\label{sec:algorithm_summary}
%To provide a comprehensive overview of the proposed MR-PHE framework, we present Algorithm~\ref{alg:MR-PHE}, which outlines the key steps involved in zero-shot histopathology image classification using multi-resolution prompt-guided hybrid embeddings. This algorithm integrates the processes described in the previous subsections, including multi-resolution patch extraction, hybrid embedding construction, prompt generation and selection, and final prediction.

\section{Experiments and Results}
\label{sec:experiments}

We evaluated the effectiveness of our proposed MR-PHE framework in zero-shot visual classification of histopathology images from six datasets. Comparisons were made with various existing ZSL approaches commonly used in the natural image domain, as well as with several domain-specific foundation models and fully supervised models.

\subsection{Histopathology Datasets}
We evaluated our proposed MR-PHE framework on six histopathology datasets (Table~\ref{tab:datasets}): the breast histopathology dataset BRACS~\cite{brancati2022bracs}, the colorectal histopathology datasets CRC100K~\cite{kather2018crc100k} and EBHI~\cite{hu2023ebhi}, the gastric histopathology dataset HE-GHI-DS~\cite{chen2022gashis}, the lung histopathology dataset WSSS4\-LUAD~\cite{Han2022Multilayer}, and our in-house breast histopathology dataset. For binary classification, the seven classes of BRACS were grouped into noninvasive (containing NT, PB, UDH, FEA, ADH, DCIS) and invasive (IC).
%BRACS comprises images in seven categories: normal tissue (NT, \( n=484 \)), pathological benign (PB, \( n=836 \)), usual ductal hyperplasia (UDH, \( n=517 \)), flat epithelial atypia (FEA, \( n=756 \)), atypical ductal hyperplasia (ADH, \( n=507 \)), ductal carcinoma in situ (DCIS, \( n=790 \)), and invasive carcinoma (IC, \( n=649 \)). For binary classification, we grouped these classes into non-invasive (NT, PB, UDH, FEA, ADH, DCIS) and invasive (IC) categories. Our in-house dataset consists of benign (\( n=3,138 \)) and malignant (\( n=9,018 \)) images,  ensuring high-quality annotations and diverse morphological representations. CRC100K consists of nine classes: adipose (\( n=1,338 \)), background (\( n=847 \)), debris (\( n=339 \)), lymphocytes (\( n=634 \)), mucus (\( n=1,035 \)), smooth muscle (\( n=592 \)), normal colon mucosa (\( n=741 \)), cancer-associated stroma (\( n=421 \)), and colorectal adenocarcinoma epithelium (\( n=1,233 \)). EBHI is divided into benign (\( n=885 \)) and malignant (\( n=870 \)) classes. HE-GHI-DS is classified as normal (\( n=140 \)) and abnormal (\( n=560 \)). Finally, WSSS4LUAD includes images categorized into three classes: tumor (\( n=6,579 \)), stroma (\( n=7,076 \)), and normal (\( n=1,832 \)).

\begin{table}[!t]
\centering
\caption{Histopathology datasets used in the experiments. For each dataset the number of classes and the number of images in each class are listed.}
\label{tab:datasets}
\begin{tabular}{llc}
\toprule
\textbf{Dataset} & \textbf{Classes} & \textbf{Images} \\
\midrule
& Normal Tissue (NT) & 484 \\
& Pathological Benign (PB) & 836 \\
& Usual Ductal Hyperplasia (UDH) & 517 \\
BRACS & Flat Epithelial Atypia (FEA) & 756 \\
& Atypical Ductal Hyperplasia (ADH) & 507 \\
& Ductal Carcinoma In Situ (DCIS) & 790 \\
& Invasive Carcinoma (IC) & 649 \\
\midrule
& Adipose & 1,338 \\
& Background & 847 \\
& Debris & 339 \\
& Lymphocytes & 634 \\
CRC100K & Mucus & 1,035 \\
& Smooth Muscle & 592 \\
& Normal Colon Mucosa & 741 \\
& Cancer-Associated Stroma & 421 \\
& Colorectal Adenocarcinoma Epithelium & 1,233 \\
\midrule
\multirow{2}*{EBHI}
& Benign & 885 \\
& Malignant & 870 \\
\midrule
\multirow{2}*{HE-GHI-DS}
& Normal & 140 \\
& Abnormal & 560 \\
\midrule
& Tumor & 1,181 \\
WSSS4LUAD & Stroma & 1,680 \\
& Normal & 1,832 \\
\midrule
\multirow{2}*{In-House}
& Benign & 3,138 \\
& Malignant & 9,018 \\
\bottomrule
\end{tabular}
\end{table}

\begin{table*}[!t]
\centering
\scriptsize
\caption{Performance comparison of MR-PHE against various ZSL methods on multiple histopathology datasets. All methods use the same pretrained image and text encoders from the CONCH model for a fair comparison, differing only in prompt engineering or weighting strategies. The best performance for each dataset is highlighted in \textbf{bold} and the second-best is \underline{underlined}. The last column ($\Delta$) shows the improvement of MR-PHE over the second-best method.}
\label{tab:results}
\setlength{\tabcolsep}{4pt}
\resizebox{\textwidth}{!}{%
\begin{tabular}{llcccccccc}
\toprule
\textbf{Dataset} & \textbf{Metric} 
& \textbf{CLIP} & \textbf{CLIP-D} & \textbf{CLIP-E} 
& \textbf{CuPL} & \textbf{Waffle} & \textbf{WCA} 
& \textbf{MR-PHE} & \(\Delta\) \\
\midrule

\multirow{2}{*}{BRACS (7-Class, \(n = 4{,}539\))} 
 & F1 (\%)  
   & 28.5 & 24.04 & 34.0 & \underline{44.09} & 31.88 & 43.87 
   & \textbf{51.63} & 7.54 \\
 & Acc (\%) 
   & 33.4 & 28.24 & 37.08 & \underline{53.54} & 36.53 & 48.46 
   & \textbf{58.93} & 5.39 \\
\midrule

\multirow{2}{*}{BRACS (2-Class, \(n = 4{,}539\))} 
 & F1 (\%)  
   & 62.16 & 24.47 & \underline{78.54} & 78.20 & 63.90 & 70.29 
   & \textbf{90.36} & 11.82 \\
 & Acc (\%) 
   & 84.21 & 24.48 & 86.55 & 85.49 & \underline{87.33} & 79.28 
   & \textbf{95.80} & 8.47 \\
\midrule

\multirow{2}{*}{CRC100K (\(n = 7{,}180\))} 
 & F1 (\%)  
   & 67.97 & 56.76 & 58.24 & 59.97 & \underline{68.34} & 30.87 
   & \textbf{76.95} & 8.61 \\
 & Acc (\%) 
   & 74.21 & 62.84 & 65.64 & 62.62 & \underline{75.13} & 31.80 
   & \textbf{80.22} & 5.09 \\
\midrule

\multirow{2}{*}{EBHI (\(n = 1{,}755\))} 
 & F1 (\%)  
   & 90.65 & 89.0 & \underline{92.18} & 74.40 & 89.77 & 87.36 
   & \textbf{94.46} & 2.28 \\
 & Acc (\%) 
   & 90.66 & 89.0 & \underline{92.19} & 75.95 & 89.80 & 87.46 
   & \textbf{94.47} & 2.28 \\
\midrule

\multirow{2}{*}{HE-GHI-DS (\(n = 700\))} 
 & F1 (\%)  
   & 28.63 & 65.97 & 33.74 & \underline{91.12} & 34.80 & 82.18 
   & \textbf{95.16} & 4.04 \\
 & Acc (\%) 
   & 29.43 & 68.86 & 33.86 & \underline{94.86} & 34.86 & 86.00 
   & \textbf{96.71} & 1.85 \\
\midrule

\multirow{2}{*}{WSSS4LUAD ($n$ = 4,693)} & F1-score (\%) & 55.08 & 69.86 & 52.22 & 67.30 & 35.16 & \underline{71.91} & \textbf{89.56} & 17.65 \\
 & Accuracy (\%) & 57.45 & 69.49 & 52.55 & 67.68 & 39.68 & \underline{70.96} & \textbf{88.96} & 18.00 \\

 \midrule

\multirow{2}{*}{In-House (\(n = 12{,}156\))} 
 & F1 (\%)  
   & 69.54 & \underline{85.48} & 71.73 & 79.92 & 37.39 & 81.93 
   & \textbf{91.47} & 6.0 \\
 & Acc (\%) 
   & 71.07 & \underline{89.22} & 73.12 & 83.70 & 38.22 & 84.93 
   & \textbf{93.26} & 4.04 \\
\bottomrule
\end{tabular}
}
\end{table*}

\subsection{Implementation Details}
Our MR-PHE framework was implemented using Python 3.12.3 and PyTorch 2.3.0. All experiments were conducted on the NCI Gadi supercomputer, utilizing nodes equipped with NVIDIA DGX A100 GPUs (80 GB memory). We employed the pretrained CONCH model~\cite{lu2024visual} for both the image encoder \( f \) and text encoder \( g \) without any fine-tuning, operating in a ZSL setting. Input images were processed according to the CONCH model's configuration. Multiresolution patches were extracted using scaling factors \( \mathcal{S} = \{ 0.25, 0.50, 0.75 \} \), with 5 random patches per scale, resulting in 15 patches per image, plus the original image to incorporate global context. Key hyperparameters were set as follows: hybrid weight \( \alpha = 0.5 \), temperature \( \tau = 1.5 \), text scaling factor \( \beta = 2.0 \), and top \( K = 30 \) prompts per class selected based on validation performance. Prompts were generated using templates and clinically relevant descriptors.

\begin{table}[!t]
\centering
\footnotesize
\caption{Performance comparison of MR-PHE with histopathology domain-specific foundation models on the CRC100K and WSSS4LUAD datasets. Models marked with an asterisk ($^*$) represent results with prompt ensembling. The best performance for each dataset is highlighted in \textbf{bold} and the second-best is \underline{underlined}.}
\label{tab:domain_models_comparison}
%\begin{tabularx}{\columnwidth}{l l c c}
\begin{tabular}{llcc}
\toprule
\textbf{Dataset} & \textbf{Model} & \textbf{F1 (\%)} & \textbf{Accuracy (\%)} \\
\midrule
%\multirow{4}{*}{CRC100K} 
& MR-PHE & \textbf{76.95} & \textbf{80.22} \\
& CONCH & 59.0 & 59.8 \\
& CONCH$^*$ & \underline{70.5} & \underline{71.9} \\
CRC100K & PLIP & 40.8 & 46.2 \\
& PLIP$^*$ & 63.6 & 62.4 \\
& BiomedCLIP & 45.2 & 51.2 \\
& BiomedCLIP$^*$ & 57.5 & 61.6 \\
\midrule
%\multirow{4}{*}{WSSS4LUAD} 
& MR-PHE & \textbf{89.6} & \textbf{88.9} \\
& CONCH & \underline{86.5} & \underline{86.7} \\
& CONCH$^*$ & 57.0 & 62.0 \\
WSSS4LUAD & PLIP & 78.3 & 78.7 \\
& PLIP$^*$ & 41.9 & 53.0 \\
& BiomedCLIP & 82.9 & 83.3 \\
& BiomedCLIP$^*$ & 46.5 & 55.0 \\
\bottomrule
\end{tabular}
%\end{tabularx}
\end{table}

\begin{table}[!t]
\centering
%\scriptsize
\caption{Performance comparison of MR-PHE with existing supervised models on public datasets. 
We adopt accuracy (\%) as the performance metric, with values for the baseline models copied from their respective publications. 
\textbf{Bold} indicates the highest accuracy per dataset, while \underline{underline} indicates the second highest.}
\label{tab:comparison}
%\resizebox{\columnwidth}{!}{%
\begin{tabular}{llclc}
\toprule
\textbf{Dataset} & \textbf{Classification} & \textbf{Reference} 
& \textbf{Model} & \textbf{Accuracy (\%)} \\
\midrule

% ------------------ BRACS DATASET ------------------
\multirow{6}{*}{BRACS} 
& \multirow{4}{*}{7-Class} 
  & \cite{pati2022hierarchical} & HACT-Net 
  & \textbf{61.53$\pm$0.87} \\ 
& & \cite{brancati2022bracs} & DL-Based Method 
  & 55.90 \\
& & Ours & MR-PHE 
  & \underline{58.93} \\ 

\cmidrule{2-5}

& \multirow{2}{*}{Binary} 
  & \cite{pati2022hierarchical} & HACT-Net 
  & \textbf{96.32$\pm$0.64} \\ 
& & Ours & MR-PHE 
  & \underline{95.80} \\ 

\midrule

% ------------------ CRC100K ------------------
\multirow{4}{*}{CRC100K} 
& \multirow{4}{*}{Multiclass} 
  & \cite{wang2022transformer} & CTransPath 
  & 61.9 \\
& & \cite{he2016deep} & ResNet50 
  & 40.2 \\
& & \cite{Riasatian2021FineTuning} & KimiaNet 
  & 39.8 \\
& & Ours & MR-PHE 
  & \textbf{80.22} \\ 

\midrule

% ------------------ EBHI ------------------
\multirow{2}{*}{EBHI} 
& \multirow{2}{*}{Binary} 
  & \cite{hu2023ebhi} & ResNet50 
  & \underline{83.81} \\ 
& & Ours & MR-PHE 
  & \textbf{94.47} \\ 

\midrule

% ------------------ HE-GHI-DS ------------------
\multirow{2}{*}{HE-GHI-DS} 
& \multirow{2}{*}{Binary} & \cite{fan2023cam} & CAM-VT 
  & \underline{90.48} \\
& & Ours & MR-PHE 
  & \textbf{96.71} \\ 

\midrule

% ------------------ WSSS4LUAD ------------------
\multirow{11}{*}{WSSS4LUAD} 
& \multirow{11}{*}{Multiclass} 
  & \cite{he2016deep} & ResNet50 
  & 75.2 \\
& & \cite{huang2017densely} & DenseNet121 
  & 77.7 \\
& & \cite{tan2019efficientnet} & EfficientNet-B3 
  & 78.0 \\
& & \cite{tan2019efficientnet} & EfficientNet-B5 
  & 79.7 \\
& & \cite{liu2022convnet} & ConvNeXt-B 
  & 83.2 \\
& & \cite{liu2022convnet} & ConvNeXt-XL 
  & 84.2 \\
& & \cite{dosovitskiy2020image} & ViT-B16 
  & 70.4 \\
& & \cite{caron2021emerging} & DINO-ViT-S16 
  & 83.9 \\
& & \cite{alfasly2024rotation} & PathDINO-224 
  & 86.9 \\
& & \cite{alfasly2024rotation} & PathDINO-512 
  & 86.7 \\
& & Ours & MR-PHE 
  & \textbf{88.96} \\

\bottomrule
\end{tabular}%}
\end{table}

\begin{table}[!t]
\centering
\caption{Ablation study on the HE-GHI-DS dataset, demonstrating how each module of MR-PHE contributes to ZSL classification performance. 
Columns indicate whether each component is enabled (\checkmark) or disabled (\(\times\)). 
We report accuracy and the drop (\(\Delta\)) relative to the full MR-PHE configuration.}
\label{tab:ablation_study}
\resizebox{\columnwidth}{!}{%
\begin{tabular}{lcccccc}
\toprule
\textbf{Configuration} 
& \textbf{Multiresolution}
& \textbf{Hybrid} 
& \textbf{Patch}
& \textbf{Prompt Generation} 
& \textbf{Accuracy} 
& \(\Delta\) \\
 & \bf Patches & \bf Embedding & \bf Weighting & \bf \& Selection & (\%) & (\%) \\
\midrule
Basic Prompts          
 & \checkmark & \checkmark & \checkmark & \(\times\)  & 87.86   & -8.85 \\

Single Scale           
 & \(\times\) & \checkmark & \checkmark & \checkmark & 95.76   & -0.95 \\

Only Global Embeddings 
 & \checkmark & \(\times\) & \checkmark & \checkmark & 96.29   & -0.42 \\

Only Patch Embeddings  
 & \checkmark & \(\times\) & \checkmark & \checkmark & 96.43   & -0.28 \\

Average Patch Weighting
 & \checkmark & \checkmark & \(\times\) & \checkmark & 96.51   & -0.20 \\
\midrule
\textbf{Full MR-PHE (Proposed)} 
 & \checkmark & \checkmark & \checkmark & \checkmark & \textbf{96.71} & -- \\
\bottomrule
\end{tabular}
}
\end{table}

\subsection{Comparison With ZSL Methods}
First we compared MR-PHE against several existing ZSL approaches commonly used in the natural image domain. Specifically, we considered CLIP~\cite{clip2021radford}, a vision-language model utilizing manually created prompt templates such as ``a photo of a [CLASS]'' for zero-shot classification. We included CLIP-E~\cite{clip2021radford}, an ensemble version employing a variety of manually crafted templates to enhance performance through prompt ensembling. CLIP-D~\cite{menon2022visual} extends CLIP by leveraging large language models (LLMs) to generate descriptive prompts for each class, aiming to improve the quality of textual representations. CuPL~\cite{pratt2023what} generates higher-quality LLM-based descriptions than CLIP-D, resulting in improved zero-shot classification performance. Waffle~\cite{roth2023waffling} replaces LLM-generated descriptions with randomly generated character and word sequences to test the robustness of VLMs to nonsensical prompts. Additionally, WCA~\cite{li2024visual} enhances zero-shot classification by weighting the alignment between image and text embeddings based on their relevance, improving discrimination among classes. 
Important to note is that, for a fair comparison, we implemented all these ZSL methods using the same pretrained image and text encoders from CONCH~\cite{lu2024visual}. Hence, the differences among these approaches lie solely in their prompt generation or weighting strategies, rather than the underlying model.
We adapted the prompts of these methods to fit the histopathology domain, replacing generic terms like ``a photo of'' with domain-specific phrases such as ``a histopathology image of'' or ``a microscopic image showing'' to better capture the characteristics of histopathological images.
%The classification performance of MR-PHE and the comparative methods was evaluated on six histopathology datasets: BRACS (Binary and 7-Class), CRC100K, EBHI, HE-GHI-DS, WSSS4LUAD, and our in-house dataset. The results are summarized in Table~\ref{tab:results}.

%MR-PHE consistently outperformed all comparative methods across multiple histopathology datasets and classification tasks (Table~\ref{tab:results}). On the BRACS 7-class dataset, MR-PHE achieved an F1-score of 55.05\% and accuracy of 61.58\%, surpassing CuPL by 10.34\% and 8.79\%. In the BRACS binary task, it attained an F1-score of 83.30\% and accuracy of 83.76\%, outperforming WCA by 2.89\% and 2.88\%. On the CRC100K dataset with nine colorectal cancer tissue types, MR-PHE achieved an F1-score of 77.21\% and accuracy of 80.31\%, significantly outperforming CuPL by 17.89\% and 12.52\%, demonstrating its effectiveness in complex multiclass tasks. On the EBHI dataset, MR-PHE achieved an F1-score and accuracy of 95.09\% and 95.10\%, improving over CuPL by 1.94\%. It also attained near-perfect performance on the HE-GHI-DS dataset, with an F1-score of 99.33\% and accuracy of 99.57\%, exceeding CuPL by 1.37\% and 0.86\%. On the WSSS4LUAD dataset, MR-PHE achieved an F1-score of 89.56\% and accuracy of 88.96\%, improving over WCA by 17.65\% and 18.00\%. In our in-house dataset, MR-PHE reached an F1-score of 92.16\% and accuracy of 93.85\%, surpassing CLIP-D by 5.54\% and 4.50\%.

MR-PHE consistently outperformed all comparative methods across multiple histopathology datasets and classification tasks (Table~\ref{tab:results}), with improvements of 2.28\%--17.65\% in terms of F1-score, and 1.85\%--18.0\% in terms of accuracy. These consistent improvements across diverse datasets highlight the ability of MR-PHE to handle both binary and multiclass classification tasks involving complex tissue structures. Our domain-specific prompt engineering and hybrid embedding strategy effectively capture intricate morphological features, enhancing the alignment between visual and textual representations and leading to more accurate zero-shot classifications. These results validate our method's superiority in the histopathology domain, demonstrating its potential to aid pathologists without the need for extensive labeled datasets.

\subsection{Comparison With Domain-Specific Foundation Models}
Next we evaluated the effectiveness of MR-PHE by comparing its performance against state-of-the-art histopathology domain foundation models on two datasets: CRC100K and WSSS4LUAD. Specifically, we considered CONCH~\cite{lu2023towards}, PLIP~\cite{huang2023visual}, and BiomedCLIP~\cite{zhang2023large}, and used the F1-scores and accuracies of these models on the two datasets as reported in the original cited publications. The results (Table~\ref{tab:domain_models_comparison}) show that MR-PHE outperformed the compared models by a considerable margin, yielding 6.45\% higher F1-score and 8.32\% higher accuracy compared to the second-best model on the CRC100K dataset, and 3.1\% higher F1-score and 2.2\% higher accuracy on the WSSS4LUAD dataset.
%The F1-scores and accuracies for CONCH, PLIP, and BiomedCLIP are reported from their respective publications.

\subsection{Comparison With Fully Supervised Models}
Finally we assessed the competitiveness of MR-PHE against state-of-the-art fully supervised learning models with published results on the public histopathology datasets. %Table~\ref{tab:comparison} presents the accuracy achieved by MR-PHE and several supervised models on the BRACS, CRC100K, EBHI, HE-GHI-DS, and WSSS4LUAD datasets. 
The results (Table~\ref{tab:comparison}) demonstrate that our proposed framework performs comparable to, and often surpassing, fully supervised models without the need for any task-specific training or labeled data. This highlights the potential of our approach to facilitate histopathological image analysis in settings where labeled data is scarce or unavailable.

%On the challenging \textbf{CRC100K} dataset, which involves multi-class classification of colorectal histology images, MR-PHE achieves an accuracy of $80.31\%$, significantly outperforming existing supervised methods such as CTransPath~\cite{wang2022transformer} ($61.9\%$), ResNet50~\cite{he2016deep} ($40.2\%$), and KimiaNet~\cite{Riasatian2021FineTuning} ($39.8\%$). This substantial improvement highlights MR-PHE's effectiveness in handling complex multi-class tasks even without task-specific training.

\subsection{Ablation Study}
\label{subsec:ablation_study}

We conducted an ablation study on the HE-GHI-DS dataset (Table~\ref{tab:ablation_study}) to assess the relative impact of multiresolution patches, hybrid embeddings, patch weighting, and prompt generation and selection. Individually removing or modifying these components confirms that each contributes to the model’s overall zero-shot classification performance. Notably, the synergy between multiresolution patches, hybrid embeddings, patch weighting, and prompt generation and selection yields the highest accuracy (96.71\%), demonstrating the effectiveness of our proposed MR-PHE pipeline.

\section{Discussion}
\label{sec:discussion}

In this paper, we introduced the MR-PHE framework, a novel approach for ZSL in histopathology image classification. Our method leverages multiresolution patch extraction to mimic the diagnostic process of pathologists, capturing both fine-grained cellular details and broader tissue structures. By integrating global image embeddings with weighted patch embeddings through a hybrid embedding strategy, MR-PHE effectively combines local and global contextual information. Additionally, we developed a comprehensive prompt generation and selection framework that enriches class descriptions with domain-specific synonyms and clinically relevant features, enhancing the semantic alignment between visual and textual representations. Our similarity-based patch weighting mechanism further emphasizes diagnostically important regions during classification. Operating without the need for task-specific training or labeled data, MR-PHE offers scalability and reduces dependence on large annotated datasets, addressing key challenges in computational pathology.

Our experimental results demonstrate superior performance of MR-PHE across a wide variety of histopathology datasets and classification tasks. Specifically, MR-PHE consistently outperforms existing ZSL approaches, achieving notable improvements in both F1-score and accuracy (Table~\ref{tab:results}), highlighting the model's effectiveness in handling both binary and complex multiclass classification tasks with diverse tissue types, even without task-specific training. Similarly, MR-PHE outperformed domain-specific foundation models (Table~\ref{tab:domain_models_comparison}), despite these models being pretrained on large histopathology datasets, indicating the effectiveness of our approach in leveraging VLMs without additional domain-specific pretraining. Furthermore, MR-PHE often matches or exceeds the performance of fully supervised models (Table~\ref{tab:comparison}), underscoring the potential of MR-PHE to deliver high-quality diagnostic insights without the extensive labeled data required by traditional supervised learning approaches.

Despite the promising results, our work has several limitations, leaving avenues for future research. One limitation is the reliance on specific hyperparameter settings, such as the hybrid weight $\alpha$, temperature $\tau$, and the number of top prompts $K$. While we have empirically selected these parameters, a systematic exploration or optimization could further enhance performance. Additionally, our framework utilizes the CONCH model as the underlying VLM. Exploring other foundation models, such as PLIP or BiomedCLIP, or ensemble approaches could potentially improve the generalization capabilities of MR-PHE. Practical deployment of our framework in clinical settings also presents challenges. While our method reduces the need for labeled data, it still requires careful prompt engineering and validation to ensure clinical relevance and accuracy. Future research should investigate domain adaptation techniques to better tailor the model to specific clinical contexts. Incorporating strategies to handle data variability, such as differences in staining techniques, imaging protocols, and tissue preparation, is also essential to enhance robustness. Additionally, integrating MR-PHE into clinical workflows as a decision-support tool could prioritize the diagnosis of malignant cases, allowing for faster reporting in high-volume settings and facilitating early intervention. Addressing challenges such as model interpretability, ensuring that clinicians can understand and trust the model's decisions, will be crucial for adoption. Collaboration with clinical experts is imperative to refine the system, ensuring it aligns with clinical needs and enhances existing workflows without introducing unnecessary complexity.

Notwithstanding its current limitations, MR-PHE is already showing promising results. By focusing on scalability, adaptability, and clinical relevance, the framework has the potential to bridge the gap between research advancements and practical applicability, ultimately improving patient care through enhanced diagnostic accuracy and efficiency.

\section*{Ethical Approval}
Ethical approval for the use of breast cancer WSIs in our in-house dataset was provided by the Human Research Ethics Committee at Prince of Wales Hospital, Sydney, with approval number HREC/17/POWH/389-17/176. This study was conducted in accordance with the guidelines of the committee and with respect for the subjects' data.

\vspace{-1.5\baselineskip}
\bibliographystyle{unsrt} % IEEEtran bibstyle
\bibliography{paper} % your .bi

\end{document}